\documentclass[letterpaper, 10 pt, conference]{ieeeconf}  

\IEEEoverridecommandlockouts                              

\usepackage{times}

\usepackage[dvipsnames,svgnames,table]{xcolor}
\usepackage{multicol}
\usepackage[bookmarks=true]{hyperref}
\usepackage{amsmath,amssymb,amsfonts,mathtools,bm}
\usepackage{accents}
\usepackage{comment}
\usepackage[normalem]{ulem}
\usepackage{graphicx}
\usepackage{float}
\usepackage{balance}
\usepackage{stmaryrd}
\usepackage{rotating}
\usepackage{subcaption}
\usepackage{tablefootnote}

\usepackage{tabularx}
\usepackage{makecell}

\usepackage{graphicx}
\usepackage{subcaption}

\usepackage{algorithm}
\usepackage{algpseudocode}

\usepackage{cuted}

\usepackage{tikz}
\usetikzlibrary{positioning,arrows.meta,shadows.blur,calc}
\usepackage{fontawesome5}

\usepackage{adjustbox}   

\newcommand{\bfm}{\mathbf}

\title{\LARGE \bf
Learning from Demonstrations over Riemannian Manifolds using Neural ODEs: An Extended Abstract
}

\author{Diana Cuervo Espinosa$^{1}$, Mahathi Anand$^{2}$ and Angela P. Schoellig$^{2}$
\thanks{*This work is supported in part by Robotics Institute Germany, funded by BMFTR grant 16ME0997K}
\thanks{$^{1}$Chair of Robotics and System Intelligence, Technical University of Munich, Germany
\tt{diana.cuervo@tum.de}        }%
\thanks{$^{2}$Learning Systems and Robotics Lab, Technical University of Munich, Germany
  \tt{\{mahathi.anand, angela.schoellig\}@tum.de} }%
  }

\begin{document}

\maketitle
\thispagestyle{empty}
\pagestyle{empty}

\begin{abstract}

Learning from demonstratins (LfD) is usually performed over Euclidean spaces, while the robot state, e.g. orientation, naturally evolves over curved spaces. Therefore, to ensure natural, complex motion generation, we investigate learning from demonstrations over Riemannian manifolds that are capable of encoding both position and orientation data. Here, geodesic paths provide for natural motion between two arbitrary points within the manifold. We propose to numerically estimate geodesics via neural ordinary differential equations, mitigating large computational overhead of existing approaches. Finally, these geodesics can be decoded back into the original task space before deploying on the robot. In this extended abstract, we discuss the architecture of our framework, provide some initial insights from our simulation experiments, including comparison to other geodesic computation mechanisms, and discuss the challenges and prospects for future work. 
\end{abstract}

\section{MOTIVATION}

Traditional methods for robot motion planning are usually hard coded for specific tasks and require extensive coding expertise. This makes it difficult to adapt to new tasks, and significant time and effort is required to identify relevant goal locations or sequential waypoints. Learning from demonstrations seeks to enhance generalization by collecting demonstrations from human experts and directly converting them to robot motion in an adaptive manner without requiring significant programming experience~\cite{BillardMirrazaviFigueroa2022book, saveriano_dynamic_2023}. However, they are usually limited to generating motion in the Euclidean task space. On the other hand, robot orientations, which evolve over curved manifolds, cannot be represented accurately with a purely Euclidean framework, and any extension to capturing full end-effector data requires careful enforcement of geometric constraints~\cite{ravichandar_learning_2019}. 

Recently, a Riemannian perspective to LfD has emerged~\cite{beik-mohammadi_learning_2021}. Here, demonstrations are encoded into a curved latent space, i.e. a Riemannian manifold, where full end-effector poses can be described naturally. Then, robot motion may be learned directly via geodesics, i.e., the shortest paths between two arbitrary points. However, finding geodesics is a challenging problem, as it requires solving complex, second order differential equations. Several numerical relaxations, including iterative schemes~\cite{arvanitidis_fast_2019} and graph-based methods~\cite{beik-mohammadi_learning_2021} are common. Nevertheless, they are either computationally intensive, or rely on discrete approximations making it tedious to interpolate motion between two unseen data points. In this work, we propose an adaptive motion generation pipeline over Riemannian manifolds by utilizing variational autoencoders for encoding demonstrations into manifolds and neural ordinary differential equations (NODE) for geodesic computation. The NODE can quickly generate a motion path between any two locations in the manifold at inference time, making our approach very attractive for adaptive generalization and for hybrid, sequential tasks.  

\begin{figure}[t!]
\centering
\includegraphics[width = 0.48\textwidth]{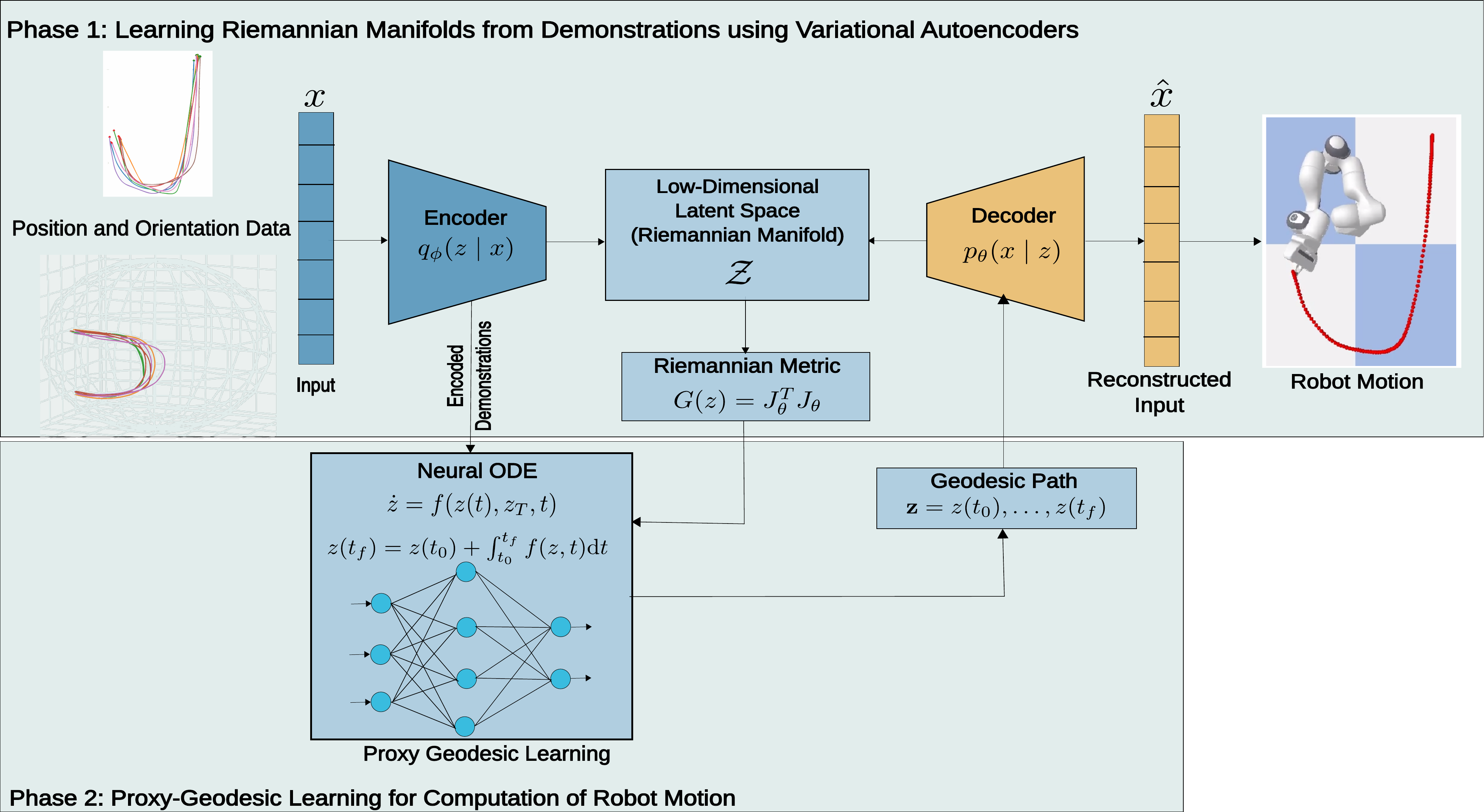}
\caption{The architecture for learning from demonstrations over Riemannian manifolds.}
\label{fig:pipeline}
\vspace{-2em}
\end{figure}

\section{ARCHITECTURE}
~\label{sec:arch}
The proposed Riemannian motion generation framework is decoupled into two phases as described in Fig.~\ref{fig:pipeline} -- the learning of robot's spatial constraints via the Riemannian manifold, and solving for the robot's temporal dynamics via geodesic paths in the learned manifold. The first phase is similar to the one proposed in~\cite{beik-mohammadi_learning_2021}, where a variational autoencoder (VAE) takes the demonstration data $x \in \mathcal{X}$ as input and encodes it into a lower-dimensional latent variable $z \in \mathcal{Z}$. In particular, the variational autoencoder consists of an encoder with parameters $\phi: \mathcal{X} \to \mathcal{Z}$, responsible for approximating the posterior distribution $p_\phi(z \mid x)$, and a decoder with parameters $\theta: \mathcal{Z} \to \mathcal{X}$ that approximates the generative distribution $p_\theta(x \mid z)$. It is trained under the standard evidence lower bound (ELBO) loss that consists of a regularization term as well as a reconstruction term. The resulting decoder function $\theta$ is then used to construct a local Riemannian metric given by $M(z) = J_\theta^T (z) J_\theta(z), \forall z \in \mathcal{Z}$, which characterizes the length and energy of a curve $\bfm z : [0,1] \to \mathcal{Z}$ as
$l_\bfm{z} = \int_{0}^1 \sqrt{\dot{\bfm z}(t)^T M(\bfm z(t)) \dot{\bfm z}(t)} \mathrm{d}t,$ and $
E_{\bfm z} = \frac{1}{2} \int_{0}^1  \dot{\bfm z}(t)^T M(\bfm z(t)) \dot{\bfm z}(t)\mathrm{d}t$, respectively. As a result, finding the geodesic, i.e., the shortest path between any two points in the manifold is reduced to finding a path that minimizes the energy (or length), respectively. 

Having learned a suitable manifold capturing the demonstrations, one then needs to find a geodesic path between two arbitrary points. Knowing that the computation of geodesics naturally reduces to solving a differential equation, we utilize NODEs to compute proxy-geodesics, i.e., geodesic paths that are estimated by utilizing the learned manifold and the data from the demonstrations. Specifically, we use a goal-parameterized NODE given by $\frac{\mathrm{d} \bfm z}{\mathrm{d}t} = f_\psi(\bfm z(t), \bfm z_g, t)$, where $f_\psi$ is a standard feedforward neural network parameterized by $\psi$. To ensure smoothness of $f_\psi$, we use smooth neural activation functions.  Then, the NODE is trained to minimize a loss function that takes into account (i) the energy of the path, (ii) imitation to demonstrations, and (iii) goal-reaching. Once trained, the NODE generates a solution geodesic path for any given start-goal pair $(z_0,z_T)$ by solving a suitable initial value problem numerically. The obtained path is then decoded via the decoder function $\theta$ to obtain the full end-effector motion path in the original task space for the robot to track using a suitable low-level controller.

\section{INSIGHTS}
\label{sec:insights}
\begin{figure}
\centering
\includegraphics[scale=0.15]{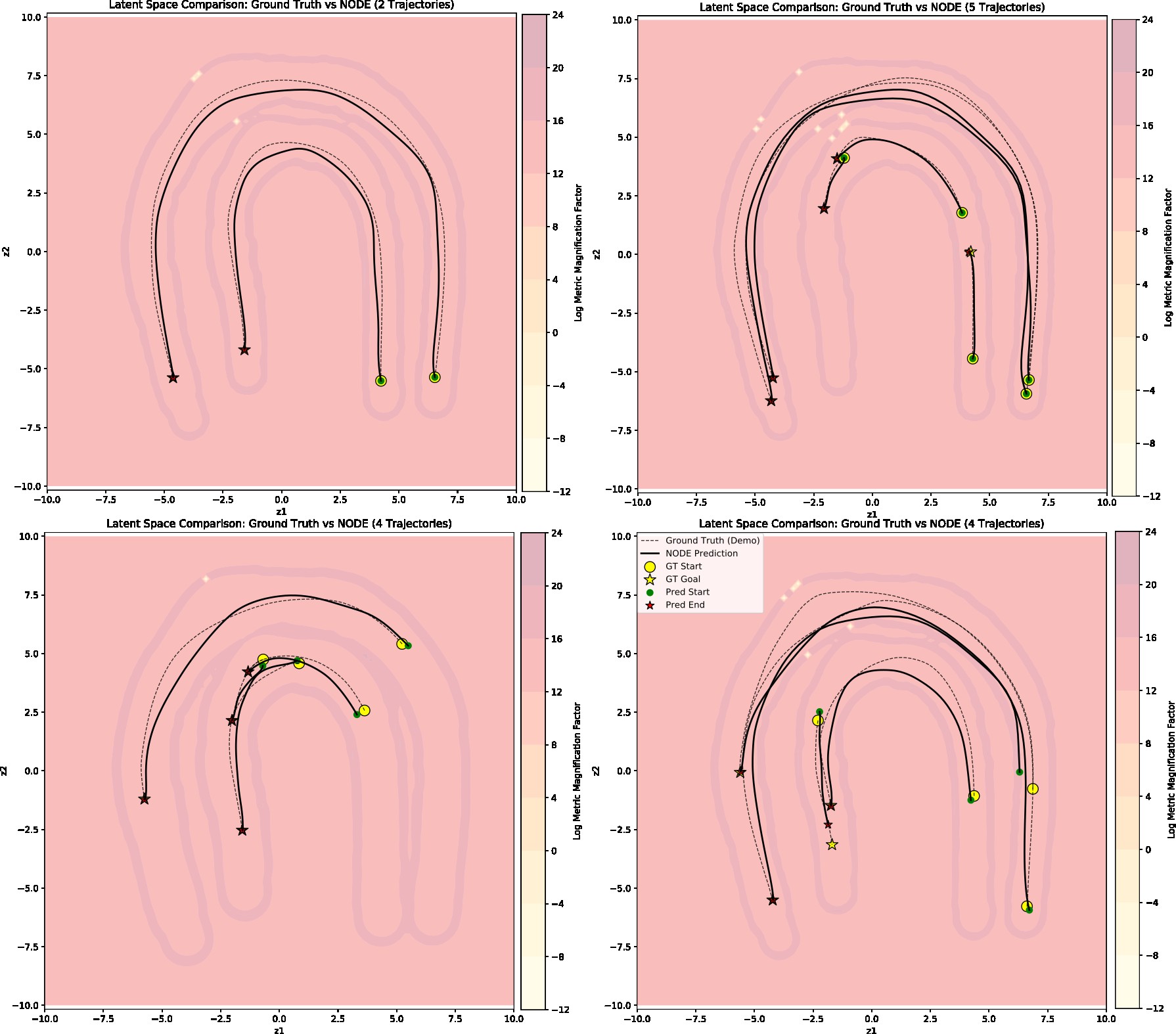}
\caption{Simulation plots illustrating the $3$ scenarios - baseline (top left), generalization (top right), and generalization with perturbation (dist. $0.02$ (bottom left) and 0.04 (bottom right)).}
\label{fig:plot}
\vspace{-2em}
\end{figure}
The results of our proposed approach were validated on a simple case study where the demonstration data~\cite{beik-mohammadi_learning_2021} is restricted to $\mathbb{R}^2 \times \mathcal{S}^2$, with the position data resembling a $J$-shape and orientation data following a $C$-shape projected on a $3D$ sphere as in Fig.~\ref{fig:pipeline}. Note that we used the same VAE as the one trained in~\cite{beik-mohammadi_learning_2021}. The learned manifold can be visualized in Fig.~\ref{fig:plot}. Here, the boundary of the manifold can also be distinctly observed (darker pink) due to the low data density in the region (measured via log-magnification factor $\log \sqrt{\det M}$). Then, we train a suitable NODE as described in Section~\ref{sec:arch} and use it during inference for the computation of geodesics. We perform several simulation case studies to test the following scenarios: (i) \textbf{baseline} case to test whether the geodesics can faithfully imitate the demonstrations encoded in the latent space. Here, the evaluation was performed by picking start and goal locations directly from the test dataset, and compared against the ground truth values, (ii) \textbf{generalization} case to study the adaptivity of the paths when initialized with arbitrary start and goal locations from the test demonstrations, and (iii) \textbf{generalization with perturbation} (Gaussian) to investigate the closeness of paths from the ground truth when starting from locations slightly perturbed from the test data. The simulation results can be visualized in Fig.~\ref{fig:plot}. One notices that the learned paths tend to stay within the manifold clusters, and respect the boundary of the manifold. In addition, we evaluate several quantitative metrics, including target convergence error w.r.t. ground truth data as well as time taken to generate the proxy geodesic path between arbitrary start and goal points at inference time. These metrics are also compared against existing geodesic computation method used in \cite{beik-mohammadi_learning_2021} as shown in Tables~\ref{tab:1} and~\ref{tab:2}. The results show that our method provides competitive target convergence and adaptability, but the biggest win is in the inference time, which is performed much faster than the discrete approach presented in ~\cite{beik-mohammadi_learning_2021}.

\begin{table}[t!]
\centering
\resizebox{1\columnwidth}{!}{
\begin{tabular}{ c|cccc}
scenario & baseline & gen. & gen. (std. $0.2$) & gen. (std. $0.4$)  \\
\hline
tar. conv. error & $0.01$ & $0.08$ & $0.17$ & $0.32$ \\
inf. time (s) & $0.27$ & $8.50$& $6.54$ & $6.49$
\end{tabular}}
\caption{The evaluation metrics (average) of our method  obtained from $7000$ proxy-geodesic computations}
\label{tab:1}
\vspace*{-\baselineskip}
\end{table}

\begin{table}[h]
\centering
\begin{tabular}{l|ccc}
& \textbf{ours} & \multicolumn{2}{c}{\textbf{graph-based~\cite{beik-mohammadi_learning_2021}}} \\
& & 40$\times$40 nodes & 100$\times$100 nodes \\
\hline
inference time (s) & 3.353 & 372.310 & 426.199 \\
tar. conv. error & 0.101 & 0.1866 & 0.007 \\
\end{tabular}
\caption{Quantitative comparison when computing geodesics with our method vs. existing approaches. Here, aevrage values are computed from 3500 proxy-geodesic computations over 15 trials. Outliers in the data are excluded.}
\label{tab:2}
\end{table}

\section{CHALLENGES AND FUTURE WORK}

While we observed that the proxy-geodesic remains within the manifold empirically, the current framework provides no guarantees on the invariance within the manifold (particularly since the NODE is trained with local Euclidean loss functions and forward integration). There are also no convergence guarantees to the goal. Therefore, future work will focus on using Riemmannian geometry and formal methods to rigorously evaluate the correctness of the approach. More extensive experimentation on a real robot is also in order.

\addtolength{\textheight}{-12cm}   




\bibliographystyle{IEEEtran}
\bibliography{bibliography}

\end{document}